\documentclass[sigconf]{acmart}

\usepackage{booktabs,multirow,array,caption}

\AtBeginDocument{%
  \providecommand\BibTeX{{%
    \normalfont B\kern-0.5em{\scshape i\kern-0.25em b}\kern-0.8em\TeX}}}

\copyrightyear{2020}
\acmYear{2020}
\setcopyright{acmcopyright}\acmConference[MM '20]{Proceedings of the 28th ACM International Conference on Multimedia}{October 12--16, 2020}{Seattle, WA, USA} \acmBooktitle{Proceedings of the 28th ACM International Conference on Multimedia (MM '20), October 12--16, 2020, Seattle, WA, USA}
\acmPrice{15.00}
\acmDOI{10.1145/3394171.3413986}
\acmISBN{978-1-4503-7988-5/20/10}

\begin{document}
\fancyhead[]{}
\title{Adaptively-Accumulated Knowledge Transfer \\for Partial Domain Adaptation}


\author{Taotao Jing$^\dagger$, ~Haifeng Xia$^\dagger$, ~ Zhengming Ding$^{\sharp}$}
\affiliation{%
  \institution{$^\dagger$Department of ECE, Indiana University-Purdue University Indianapolis, USA}
    \institution{$^{\sharp}$Department of CIT, Indiana University-Purdue University Indianapolis, USA}
  \streetaddress{}
  \city{}
  \country{}}
\email{{jingt,haifxia,zd2}@iu.edu}





\begin{abstract}
Partial domain adaptation (PDA) attracts appealing attention as it deals with a realistic and challenging problem when the source domain label space substitutes the target domain. Most conventional domain adaptation (DA) efforts concentrate on learning domain-invariant features to mitigate the distribution disparity across domains. However, it is crucial to alleviate the negative influence caused by the irrelevant source domain categories explicitly for PDA. In this work, we propose an Adaptively-Accumulated Knowledge Transfer framework (A$^2$KT) to align the relevant categories across two domains for effective domain adaptation. Specifically, an adaptively-accumulated mechanism is explored to gradually filter out the most confident target samples and their corresponding source categories, promoting positive transfer with more knowledge across two domains.
Moreover, a dual distinct classifier architecture consisting of a prototype classifier and a multilayer perceptron classifier is built to capture intrinsic data distribution knowledge across domains from various perspectives. By maximizing the inter-class center-wise discrepancy and minimizing the intra-class sample-wise compactness, the proposed model is able to obtain more domain-invariant and task-specific discriminative representations of the shared categories data. Comprehensive experiments on several partial domain adaptation benchmarks demonstrate the effectiveness of our proposed model, compared with the state-of-the-art PDA methods.
\end{abstract}




\maketitle

\section{Introduction}

Deep Neural Networks (DNNs) have achieved promising performances in various multi-media applications with the help of sufficient well-labeled training data, which is not always available and dramatically expensive to collect and annotate \cite{simonyan2014very,yan2016image,he2016deep, ding2018robust}. Domain Adaptation (DA) has made significant progress in such a common and real-world situation when massive amounts of well-labeled training data of the target domain are not accessible \cite{jiang2017deep,JADA,zhuo2017deep,wang2018visual,yao2019heterogeneous,li2019cycle, Xia_2020_CVPR}. The philosophy of domain adaptation is transferring the knowledge of a related well-labeled source domain to the unlabeled target domain by aligning the marginal and conditional distributions while mitigating the domain distribution disparity across domains. Towards this goal, a plenty of domain adaptation (DA) techniques have been successfully applied in various multimedia tasks such as multimodal learning~\cite{multimodal_learning2,DTN,wang2020ev}, visual object recognition~\cite{JADA,DCAN}, and text categorization~\cite{text_categorization3,text_categorization2}. 


Recent domain adaptation efforts seek to capture general domain-invariant but task-discriminative feature representations in shared feature space for two domains through matching the cross-domain distribution alignment schemes. Discrepancy loss is one of the most commonly used strategies to evaluate the cross-domain distribution difference, e.g., maximum mean discrepancy (MMD) \cite{borgwardt2006integrating}. A bunch of domain adaptation efforts design various MMD loss functions to align the source and target domain marginal and conditional distribution by incorporating pseudo labels of the target domain \cite{DAN,RTN}. Besides, the adversarial loss is another sufficiently explored scheme to eliminate the domain shifts by training one or more domain discriminator against the feature generator in an adversarial manner \cite{bousmalis2017unsupervised,tzeng2015simultaneous,ADDA,hoffman2017cycada,luo2017label}. Moreover, latest DA research works jointly consider both the domain-wise alignment as well as the task-specific category-level alignment \cite{saito2018maximum,lee2019sliced,zhang2019domain}, or propose various reconstruction penalties to obtain target specific structures \cite{zhang2018collaborative}. However, all conventional domain adaptation solutions assume that the source and target domain have identical label space, which is not always satisfied in real life \cite{Office31}.

Partial domain adaptation (PDA) focuses on such a common and challenging situation when the source domain label space subsumes the target domain label space \cite{PADA, SAN, IWAN}. Along this line, Cao et al. propose the Partial Adversarial Domain Adaptation (PADA) to simultaneously eliminates the negative transfer by down-weighing the source domain outlier categories during training the classifier and domain adversary and promotes the cross-domain distribution alignment in the shared label space \cite{PADA}. Later on, Cao et al. extend PADA to Selective Adversarial Networks (SAN) which incorporates instance-level and category-level weighting mechanism with multi-discriminator domain adversarial networks to not just down-weigh the source outlier classes, but also align each target sample to several most relevant classes and promote positive transfer for each instance \cite{SAN}. On the other hand, Zhang et al. present Importance Weighted Adversarial Nets (IWAN) to alleviate the distraction of the source domain outlier classes by assigning importance score of each source sample obtained from the two domain classifier strategy \cite{IWAN}.  Similar to this, Cao et al. propose Example Transfer Network (ETN) to quantify the transferability of the source samples and evaluate each point contribution to both the classifier and domain discriminator \cite{ETN}. Unfortunately, even most of the previously mentioned PDA efforts explore the re-weighing mechanism to reduce the outlier source categories negative transfer, adapting the cross-domain distribution in the whole source and target domain data space and label space is still vulnerable to the outlier source categories and misclassified samples. Besides, most existing PDA methods suffer from explicitly matching source and target domains distribution by only considering the domain-wise adaptation while ignoring the alignment of class-wise distribution. 

In this paper, we propose an Adaptively-Accumulated Knowledge Transfer scheme (A$^2$KT) to manage partial domain adaptation challenges by simultaneously promoting positive transfer in the shared label space while alleviating negative transfer caused by the outlier source categories. The general idea is through gradually filtering out confident task-relevant target samples and corresponding categories to optimize both domain-wise distribution adaptation and class-wise distribution alignment. To sum up, the contributions of this paper are highlighted as follows:
\begin{itemize}
    \item First of all, we propose an adaptively-accumulated knowledge transfer strategy to iteratively weigh and filter out confident task-relevant target samples and corresponding categories under the guidance of the source domain data for effective cross-domain alignment.
    \item Secondly, we explore two different types of task-specific classifiers to capture and transfer intrinsic distribution knowledge across domains from various perspectives.
    \item Thirdly, we propose a cross-domain alignment loss function which is able to align the class-level discrimination across domains, and compact the sample-level distribution within the same class.
\end{itemize}

\section{Related Work}

\subsection{Domain Adaptation}
The cross-domain data distribution discrepancy, known as domain shift, is the main challenge of domain adaptation. Plenty of works exploits the potential of deep neural networks to capture explanatory attributes and domain-invariant features in recent years, which is conducive to mitigating domain shift while transferring underlying knowledge across domains in domain adaptation tasks \cite{bengio2013representation,donahue2014decaf,yosinski2014transferable}. Compared to traditional machine learning based domain adaptation solutions,  introducing deep architecture into domain adaptation promotes the generalization of frameworks dramatically \cite{hoffman2014lsda,oquab2014learning}.  Some researchers integrate high-order statistical properties of different domains into a unified framework, such as maximum mean discrepancy (MMD), to align the data distribution across domains, which successfully eliminate domain shift and achieve promising classification performance on the target domain \cite{DAN,RTN}. By virtue of generative adversarial techniques, some works involve a domain discriminator into the game to distinguish which domain the sample belonging to while optimizing the generator and discriminator in an adversarial manner \cite{DANN,tzeng2015simultaneous,JADA}. Moreover, the latest works rethink the domain adaptation problem from various perspectives and propose dual-classifiers-based frameworks that seek to align not only domain-wise data distributions but also classifier-class-specific boundaries \cite{MCD,SWD,zhang2019domain}. 

\vspace{-3mm}
\subsection{Partial Domain Adaptation}

Unfortunately, realistic application scenarios hardly satisfy the standard domain adaptation assumption that the source and target domains share identical label space. A more common situation is the source domain subsumes the target domain label space, which means the source domain includes samples from more different categories except for the shared ones with the target domain. Such a novel challenge, named as partial domain adaptation (PDA), attracts substantial attention in transfer learning and brings out many inspiring works on this topic. Selective Adversarial Network (SAN) explores multiple adversarial networks to weight and select out the outlier categories source samples and down their transferring weights \cite{SAN}. Partial Adversarial Domain Adaptation (PADA) extends SAN and pays more attention to class-level transferability weighting on the source classifier \cite{PADA}. Similarly, Importance Weighted Adversarial Nets (IWAN) considers the sigmoid output of an auxiliary domain classifier as the indicator to measure the probability of each source sample comes from the target domain \cite{IWAN}.  Example Transfer Network (ETN) further explains the discriminative information as the transferability quantification of the source domain samples, through which the irrelevant examples from outlier categories are down-weighted for both the task-specific classifier and domain discriminator \cite{ETN}. All the pioneering efforts achieve impressive performance improvements over conventional domain adaptation approaches on PDA tasks.

However, although most existing PDA solutions seek to mitigate the negative transfer caused by outlier source classes by re-weighting samples' importance to reduce the distraction, they still train and predict the entire source domain label space, which dilutes the contribution of discriminative information within the shared categories across domains. Besides, some of them regard the prediction of the target samples as pseudo labels to align cross-domain conditional distribution, which would involve severe classification errors and mislead the optimizing direction of the model, especially at the initial stage of training when the classifier cannot handle the differently distributed unlabeled target domain samples. 

Unlike previous efforts, our proposed Adaptively Accumulated Knowledge Transfer framework (A$^2$KT) can simultaneously align the data distribution inter-class center-wise and intra-class sample-wise, both within and across domains. Exploiting prototype classifier and adaptively optimization strategy makes for eliminating the distraction triggered by the misclassified target domain samples.

\begin{figure*}
  \includegraphics[width=0.9\linewidth]{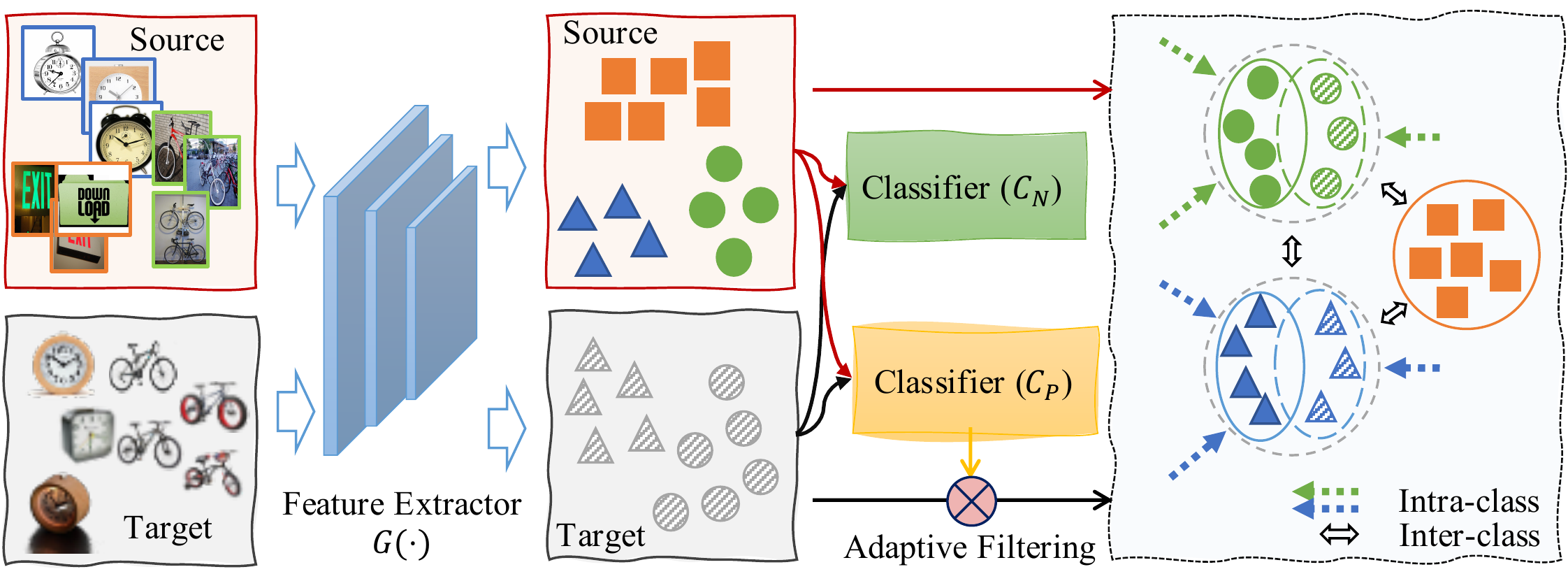}
\vspace{-3mm}
  \caption{Illustration of the proposed model for partial domain adaptation, where the source contains more categories than the target. Both source and target data are input to the feature extractor $C(\cdot)$, then classified by multilayer perceptron classifier $C_N(\cdot)$ and prototype classifier $C_P(\cdot)$. The prediction results of $C_P(\cdot)$ are exploited to filter out confident target samples for further alignment across domains. Each shape denotes one category, colored and grey shapes mean the source and target samples respectively, while the colored but shaded shapes denote the filtered out target samples with assigned pseudo labels.
  }\vspace{-2mm}
\label{fig:framework}
\end{figure*}

\section{The Proposed Method}


\subsection{Preliminaries and Motivation}

Given source domain $\mathcal{D}_s = \{\mathbf{X}_s, \mathbf{Y}_s\}=\{ (\mathbf{x}_s^1, \mathbf{y}_s^1), \cdots, (\mathbf{x}_s^{n_s}, \mathbf{y}_s^{n_s}) \}$ with labels, and unlabeled target domain $\mathcal{D}_t = \{\mathbf{X}_t\} = \{\mathbf{x}_t^1, \cdots, \mathbf{x}_t^{n_t}\}$, where $\mathbf{x}_{s/t}^{i} \in \mathbb{R}^{d}$ is a $d$-dimension source/target sample and $y_{s}^{i}$ is the known label corresponding to the source sample. $\mathcal{D}_s$ and $\mathcal{D}_t$ are drawn from distribution $P_s$ and $P_t$ respectively, while $P_s \neq P_t$. Since the source domain label space $\mathcal{C}_s$ subsumes the target domain label space $\mathcal{C}_t$, i.e., $\mathcal{C}_s \supset \mathcal{C}_t$, partial domain adaptation attempts to predict unlabeled target samples with the relevant source knowledge out of the entire well-labeled source domain. 

To eliminate the influence of irrelevant source categories, existing partial domain adaptation models mainly design a weighting strategy to select the relevant source categories for effective cross-domain alignment with discrepancy loss \cite{IWAN} or adversarial loss \cite{SAN}. To mitigate the conditional distribution mismatch across two domains, most of them rely on the pseudo labels of target samples assigned from a source-supervised neural network classifier. Due to the cross-domain distribution gap, such pseudo labels are not reliable, which would further hurt the cross-domain alignment, since the neural network classifier fits perfectly for the source distribution while not for target distribution.  

To address these issues, we consider not only to detect the irrelevant source categories to eliminate the negative influence but also select the most confident target samples during cross-domain alignment. Thus, our proposed model can adaptively select out a subset of the target domain samples that are highly affiliated to the source domain and corresponding categories to align across domains. Moreover, the prototype classifier \cite{snell2017prototypical} is adopted to annotate the target samples via source prototypes, since it can capture the intrinsic structure and semantic knowledge across source and target domain. Exploring the dual classifier architecture consisting of two different types classifiers, prototype classifier and multilayer perceptron classifer, extends the ability of the proposed model to reveal the task-specific knowledge from various perspectives.

\subsection{Adaptively-Accumulated Knowledge Transfer}

The proposed framework, which is shwon as Figure \ref{fig:framework}, consists of three modules: 1) domain-invariant feature generator $G(\cdot)$, 2) fully-connected multilayer perceptron classifier $C_N(\cdot)$, 3) prototype classifier $C_P(\cdot)$. $G(\cdot)$ takes the source and target data as input and maps them into a shared embedding space. The extracted features are denoted as $\mathbf{z}_{s/t}$ for the source and target domain respectively. 

\subsubsection{Building Diverse Source-Supervised Classifiers}

With $\mathbf{z}_{s/t}$ as input, $C_N(\cdot)$ and $C_P(\cdot)$ can assign labels from different perspectives, denoted as $\hat{\mathbf{y}}_{N}$ and $\hat{\mathbf{y}}_{P}$, respectively. $C_N(\cdot)$ is a fully-connected multilayer perceptron classifier, while the prototype classifier $C_P(\cdot)$ measures the similarity between every target sample to each source domain class center $\mu_c$ followed by a \textit{Softmax} function to assign probability prediction, that is, $\hat{\mathbf{y}}_{P,t}^{c}=\mathbf{\Phi}\big(\mathbf{z}_t,\mu_c\big)$, where $\mathbf{\Phi}(\cdot,\cdot)$ is the similarity measurement function followed by \textit{Softmax}.

In order to maintain the performance on the source domain, we keep the supervision from source and minimize the cross-entropy loss over ground truth $\mathbf{Y}_s = \{ \mathbf{y}_s^1, \cdots, \mathbf{y}_s^{n_s}  \}$ and predicted labels $\hat{\mathbf{Y}}_{N,s} = \{ \hat{\mathbf{y}}_{N,s}^1, \cdots, \hat{\mathbf{y}}_{N,s}^{n_s} \}$ from $C_N(\cdot)$ as:
\begin{equation}
\begin{aligned}
    \label{loss_y}
    \mathcal{L}_y = \frac{1}{N_s}\sum\nolimits_{i=1}^{N_s} \mathcal{L}(\mathbf{\hat{y}}_{N,s}^{i}, \mathbf{y}_s^i)
\end{aligned}
\end{equation}
As $C_P(\cdot)$ is a parameter-free classifier, so we don't need to add supervision over source domain data to $C_P(\cdot)$.

\subsubsection{Adaptively Accumulating Cross-Domain Knowledge}

Empirical Maximum Mean Discrepancy (MMD) has been verified as a promising technique to minimize the cross-domain marginal distribution difference \cite{long2017deep}. Some very recent works also adopt pseudo labels for target domain data in order to match the conditional distribution across-domain, by minimizing the distance between the source and target domain class-wise embeddings from the same category \cite{ADDA}. However, aligning all the target categories with the predicted label information is not effective since pseudo labels are not reliable especially under the PDA settings.

To alleviate the negative impact of misclassified pseudo labels to target domain samples, as well as the outlier categories from source domain label space, we propose the \textit{Adaptively-accumulated Knowledge Transfer} strategy to discard those target samples with low prediction confidences. That is, only samples with confidently predicted probability label in 
\begin{equation}
    \label{opt_adaptive}
    \begin{aligned}
    \widetilde{\mathcal{D}}_t = \{\mathbf{x}_t^i \in \mathcal{D}_t \mid \hat{\mathbf{y}}_{P,t}^{i,c} > \mathbf{p}_0 \},
    \end{aligned}
\end{equation}
are accepted to update the cross-domain alignment, where $c$ is the pseudo label of $x_t^i$, and $\hat{\mathbf{y}}_{P,t}^{i,c}$ is the probability confidence from prototype classifier $C_P(\cdot)$ of sample $\mathbf{x_t^i}$ belonging to class $c$. $\mathbf{p_0} \in [0,1]$ is the threshold. It is noteworthy that we do not need to add another hyper-parameter to tune the model, as the probability confidence measures the similarity between the target sample to the source domain, we can let the model learn $\mathbf{p}_0$ adaptively by setting it as the average of initial probability prediction produced by prototype classifier $C_P(\cdot)$ of source domain samples belonging to ground-truth class, which is
    $\mathbf{p}_0 = \frac{1}{n_s} \sum \nolimits _{\mathbf{x}_s^j \in \mathcal{D}_s} \hat{\mathbf{y}}_{P,s}^{j,c}$, 
where $c$ is the ground-truth label of source sample $\mathbf{x}_s^j$. We only explore highly-confident target samples into the cross-domain alignment. In other words, the selected target samples may not cover the whole label space, which is reasonable and acceptable.

\subsubsection{Preserving Inter-class Discrimination}

We treat the class-wise embeddings in a different way. Instead of matching the source and target domain mean embeddings from the same category, we seek to enlarge the distance between the source and target domain mean embeddings but from different classes. Specifically, we accept the $L_2$ distance to measure the distribution difference between two embeddings from two classes ($c_i, c_j$) and two domains ($d_k, d_l$):
\begin{equation}
\label{D_center}
\begin{aligned}
\mathcal{F}_{c_i,c_j,d_k,d_l}
&= \| \mu_{d_k,c_i} - \mu_{d_l,c_j} \|^2 \\ 
&= \Big\| \dfrac{1}{N_{d_k,c_i}} \sum \limits_{u=1}^{N_{d_k,c_i}} \mathbf{z}_{d_k,c_i}^{u} - \dfrac{1}{N_{d_l,c_j}} \sum \limits_{v=1}^{N_{d_l,c_j}} \mathbf{z}_{d_l,c_j}^{v}
\Big \|^2 
\end{aligned}
\end{equation}
where $\mathbf{Z} \in \mathbb{R}^{d \times (n_s+n_t)}$ denotes the embedding feature matrix composed of $\{\mathbf{z}_s^1, \cdots, \mathbf{z}_s^{n_s}\}$ and $\{\mathbf{z}_t^1, \cdots, \mathbf{z}_t^{n_t}\}$, and $\mu_{d_{k/l},c_{i/j}} \in \mathbb{R}^{d}$ denotes the class center of data from category $c_{i/j}$ domain $d_{k/l}$.


It is noteworthy that $d_k$ and $d_l$ could be the same because we also seek to maximize the class-wise distance between different categories within the same domain. On the contrary, $c_i$ and $c_j$ are always different. The integrated inter-class discriminative alignment loss term includes \textbf{TWO} parts: (1) Aligning within source/target domain (2) Aligning across domains, which is shown as Eq. \eqref{loss_inter}:
\begin{equation}
\label{loss_inter}
\begin{aligned}
\mathcal{L}_{inter} 
= &\lambda_1\Big( \sum\limits_{c=1}^{C} \sum\limits_{\substack{c'=1, \\c' \neq c}}^{C} \dfrac{\mathcal{F}_{c,c',s,s}}{C(C-1)}  +\sum\limits_{c=1}^{\hat{C}} \sum\limits_{\substack{c'=1, \\c' \neq c}}^{\hat{C}}  \dfrac{\mathcal{F}_{c,c',t,t}}{\hat{C}(\hat{C}-1)} \Big)\\
&+ \dfrac{1}{\hat{C}}\dfrac{1}{\hat{C}-1} \sum\nolimits_{c=1}^{\hat{C}} \sum\nolimits_{\substack{c'=1, \\c' \neq c}}^{\hat{C}} \mathcal{F}_{c,c',s,t},
\end{aligned}
\end{equation}
where $\lambda_1$ is a hyper-parameter to balance the contribution of within-domain and between-domain terms in $\mathcal{L}_{inter}$. It is noteworthy that here $C$ is the number of categories in the whole domain label space only when we align the inter-class discriminative distribution within source domain ($\mathcal{F}_{c,c',s,s}$), i.e., $C = \mid \mathcal{C}_s \mid$. In other situations ($\mathcal{F}_{c,c',s,t}$, $\mathcal{F}_{c,c',t,t}$), $\hat{C}$ is the number of categories of the filtered out target domain subset $\widetilde{\mathcal{D}}_t$, which may be smaller than the number of categories in the whole source domain label space, due to the \textit{Adaptively-Accumulated Knowledge Transfer} strategy we proposed to filter out target samples with high prediction confidence.


\subsubsection{Pursuing Intra-class Compactness}

Except for maximizing the inter-class distribution distance within/across domains, we also seek to pursue more intra-class compactness. Specifically, we develop an effective loss term to reduce the intra-class variation by minimizing the distance between every two samples belonging to the same category from any domains, which is shown as:
\begin{equation}
\label{D_sample}
\begin{aligned}
\mathcal{S}_{c} 
&= \dfrac{1}{N_c (N_c-1)} \sum \nolimits_{i=1}^{N_c} \sum \nolimits_{\substack{j=1 \\ j\neq i}}^{N_c} \|  \mathbf{z}^i - \mathbf{z}^j \| ^2,
\end{aligned}
\end{equation}
where 
$N_c$ is the total number of samples belonging to class $c$ from the source domain and filtered out target samples. 
Thus, we further define the total loss of all intra-class sample-wise distance as:
\begin{equation}
\label{loss_intra}
\begin{aligned}
\mathcal{L}_{intra} &=  \dfrac{\lambda_2}{C} \sum\nolimits_{c=1}^{C} \mathcal{S}_c,
\end{aligned}
\end{equation}
where $C$ is the number of categories in the source domain label space. It is noteworthy that for the target domain, we still only align those samples filtered out with high confidence to reduce the distraction of misclassification, while for samples from the source domain are always aligned over the whole label space. $\lambda_2$ is a hyper-parameter to balance the contribution of $\mathcal{L}_{intra}$.

\begin{table*}
\linespread{1.15} 
\centering 
\caption{Comparisons of Recognition Rates ($\%$) of Partial Domain Adaptation on Office-31 Dataset (ResNet-50).} 
\vspace{-3mm}
\label{Office_Res} 
\setlength{\tabcolsep}{6pt} 
\renewcommand{\arraystretch}{1} 
\begin{tabular}{c|ccccccc} 
\hline
Method & A31$\rightarrow$W10 & A31$\rightarrow$D10 & W31$\rightarrow$A10 & W31$\rightarrow$D10 & D31$\rightarrow$A10 & D31$\rightarrow$W10 & Average \\
\hline
Source Only & 75.59$\pm$1.09 & 83.44$\pm$1.12 & 84.97$\pm$0.86 & 98.09$\pm$0.74 & 83.92$\pm$0.95 & 96.27$\pm$0.85 & 87.05$\pm$0.94 \\ 
DAN \cite{DAN}& 59.32$\pm$0.49 & 61.78$\pm$0.56 & 67.64$\pm$0.29 & 90.45$\pm$0.36 & 74.95$\pm$0.67 & 73.90$\pm$0.38 & 71.34$\pm$0.46 \\ 
DANN \cite{DANN} & 73.56$\pm$0.15 & 81.53$\pm$0.23 & 86.12$\pm$0.15 & 98.73$\pm$0.20 & 82.78$\pm$0.18 & 96.27$\pm$0.26 & 86.50$\pm$0.20 \\ 
ADDA \cite{ADDA}& 75.67$\pm$0.17 & 83.41$\pm$0.17 & 84.25$\pm$0.13 & \underline{99.85}$\pm$0.12 & 83.62$\pm$0.14 & 95.38$\pm$0.23 & 87.03$\pm$0.16 \\ 
RTN \cite{RTN}& 78.98$\pm$0.55 & 77.07$\pm$0.49 & 89.46$\pm$0.37 & 85.35$\pm$0.47 & 89.25$\pm$0.39 & 93.22$\pm$0.52 & 85.56$\pm$0.47 \\ 
IWAN \cite{IWAN}& 89.15$\pm$0.37 & 90.45$\pm$0.36 & 94.26$\pm$0.25 & 99.36$\pm$0.24 & 95.62$\pm$0.29 & 99.32$\pm$0.32 & 94.69$\pm$0.31 \\
SAN  \cite{SAN}& 90.90$\pm$0.45 & 94.27$\pm$0.28 & 88.73$\pm$0.44 & 99.36$\pm$0.12 & 94.15$\pm$0.36 & 99.32$\pm$0.52 & 94.96$\pm$0.36 \\
PADA \cite{PADA}& \underline{96.54}$\pm$0.31 & 82.17$\pm$0.37 & \underline{95.41}$\pm$0.33 & \textbf{100.00}$\pm$.00 & 92.69$\pm$0.29 & 99.32$\pm$0.45 & 92.69$\pm$0.29 \\
DRCN \cite{DRCN} & 90.80 & 94.30 & 94.80 & \textbf{100.00} & 95.20 & \textbf{100.00} & 95.90 \\
ETN  \cite{ETN}& 94.52$\pm$0.20 & \underline{95.03}$\pm$0.22 & 94.64$\pm$0.24 & \textbf{100.00}$\pm$.00 & \textbf{96.21}$\pm$0.27 & \textbf{100.00}$\pm$.00   & \underline{96.73}$\pm$0.16 \\
\hline
Ours($C_N$) & 92.18$\pm$0.12 & 92.95$\pm$0.24 & \textbf{96.14}$\pm$0.23 & \textbf{100.00}$\pm$.00 & 95.92$\pm$0.32 & \textbf{100.00}$\pm$.00 & 96.20$\pm$0.15 \\
Ours($C_P$) & \textbf{97.28}$\pm$0.33 & \textbf{96.79}$\pm$0.15 & \textbf{96.14}$\pm$0.21 & \textbf{100.00}$\pm$.00 & \underline{96.13}$\pm$0.17 & \textbf{100.00}$\pm$.00 & \textbf{97.72}$\pm$0.14 \\
\hline

\end{tabular}\vspace{-3mm}
\end{table*}

\begin{table*}
\linespread{1.15} 
\centering 
\caption{Comparisons of Recognition Rates ($\%$) of Partial Domain Adaptation on Office-31 Dataset (VGG).} 
\vspace{-3mm}
\label{Office_VGG} 
\setlength{\tabcolsep}{6pt} 
\renewcommand{\arraystretch}{1} 
\begin{tabular}{c|ccccccc} 
\hline
Method & A31$\rightarrow$W10 & A31$\rightarrow$D10 & W31$\rightarrow$A10 & W31$\rightarrow$D10 & D31$\rightarrow$A10 & D31$\rightarrow$W10 & Average \\ 
\hline

Source Only & 60.34$\pm$0.84 & 76.43$\pm$0.48 & 79.12$\pm$0.54 & 99.36$\pm$0.36 & 72.96$\pm$0.56 & 97.97$\pm$0.63 & 81.03$\pm$0.57 \\
DAN \cite{DAN} & 58.78$\pm$0.43 & 54.76$\pm$0.44 & 67.29$\pm$0.20 & 92.78$\pm$0.28 & 55.42$\pm$0.56 & 85.86$\pm$0.32 & 69.15$\pm$0.37 \\ 
DANN \cite{DANN} & 50.85$\pm$0.12 & 57.96$\pm$0.20 & 62.32$\pm$0.12 & 94.27$\pm$0.16 & 51.77$\pm$0.14 & 95.23$\pm$0.24 & 68.73$\pm$0.16 \\ 
ADDA \cite{ADDA}& 53.28$\pm$0.15 & 58.78$\pm$0.12 & 63.34$\pm$0.08 & 95.36$\pm$0.08 & 50.24$\pm$0.10 & 94.33$\pm$0.18 & 69.22$\pm$0.12 \\ 
RTN \cite{ADDA}& 69.35$\pm$0.42 & 75.43$\pm$0.38 & 82.98$\pm$0.36 & 99.59$\pm$0.32 & 81.45$\pm$0.32 & 98.42$\pm$0.48 & 84.54$\pm$0.38 \\ 
IWAN \cite{IWAN}& 82.90$\pm$0.31 & \textbf{90.95}$\pm$0.33 & 93.36$\pm$0.22 & 88.53$\pm$0.16 & 89.57$\pm$0.24 & 79.75$\pm$0.26 & 87.51$\pm$0.25 \\ 
SAN \cite{SAN} & 83.39$\pm$0.36 & \underline{90.70}$\pm$0.20 & 91.85$\pm$0.35 & \textbf{100.00}$\pm$.00 & 87.16$\pm$0.23 & 99.32$\pm$0.45 & 92.07$\pm$0.27 \\  
PADA \cite{PADA} & 86.05$\pm$0.36 & 81.73$\pm$0.34 & \textbf{95.26}$\pm$0.27 & \textbf{100.00}$\pm$.00 & 93.00$\pm$0.24 & 99.42$\pm$0.24 & 92.54$\pm$0.24 \\  
ETN \cite{ETN} & 85.66$\pm$0.16 & 89.43$\pm$0.17 & 92.28$\pm$0.20 & \textbf{100.00}$\pm$.00 & \textbf{95.93}$\pm$0.23 & \textbf{100.00}$\pm$.00 & 93.88$\pm$0.13 \\  
\hline
Ours($C_N$) & \underline{88.44}$\pm$0.24 & 86.54$\pm$0.15 & 94.98$\pm$0.38 & \textbf{100.00}$\pm$.00 & \underline{94.98}$\pm$0.21 & 99.32$\pm$0.18 & \underline{94.04}$\pm$0.19 \\  
Ours($C_P$) & \textbf{90.48}$\pm$0.23 & 90.38$\pm$0.38 & \underline{95.19}$\pm$0.16 & \textbf{100.00}$\pm$.00 & 94.67$\pm$0.19 & \underline{99.66}$\pm$0.23 & \textbf{95.06}$\pm$0.20 \\  
\hline

\end{tabular}\vspace{-3mm}
\end{table*}


\subsection{Overall Objective and Optimization}

Entropy minimization regularization is adopted to eliminate the side effect caused by the uncertainty of classifiers, due to the large domain shift and samples which are hard to transfer. Especially during the early training stage, the target domain samples are easy to be assigned to wrong categories and may deteriorate the optimization procedures. We also explore the entropy minimization regularization as:
\begin{equation}
    \label{loss_em}
    \mathcal{L}_{em} = -\frac{1}{N_t} \sum\nolimits_{i=1}^{N_t}\sum\nolimits_{c=1}^{C} \hat{\mathbf{y}}_{N,t}^{i,c} \log \hat{\mathbf{y}}_{N,t}^{i,c},
\end{equation}
where $C$ is the number of categories in source domain label space, $N_t$ is the number of samples from the target domain.

To sum up, we propose our overall objective function as:
 \begin{equation}
     \label{opt_overall}
     \begin{aligned}
     \min \limits_{G,C_N} \mathcal{L}_y + \mathcal{L}_{intra} - \mathcal{L}_{inter} + \mathcal{L}_{em}.
     \end{aligned}
 \end{equation}
 
The whole framework consists of a feature generator $G(\cdot)$, a multilayer perceptron classifier $C_N(\cdot)$, and a prototype classifier $C_P(\cdot)$. As $C_P(\cdot)$ is non-parameter, so only $G(\cdot)$ and $C_N(\cdot)$ are optimized with the objective as Eq. \eqref{opt_overall}. Specifically, $\mathcal{L}_y$ is calculated on the source domain data, while $\mathcal{L}_{em}$ is based on the whole target domain. However, $\mathcal{L}_{intra}$ and $\mathcal{L}_{inter}$ are only based on the filtered out target data, as well as the corresponding source data from the same categories as the filtered target samples pseudo labels.


\section{Experiments}


\subsection{Datasets \& Implementation Details}

\textbf{Office-31} \cite{Office31} consists of more than 4,000 images from 31 categories office common objects. The dataset includes 3 different domains: Amazon, Webcam, and DSLR. Following the protocol of \cite{SAN}, 9 different partial domain adaptation tasks are explored. For each target domain, we select the 10 shared categories across Office-31 and Caltech-256 \cite{Caltech} dataset and denoted as A10, W10, and D10. The source domain data takes the whole domain data space and denoted as A31, W31, and D31.

\noindent\textbf{Office-Home} \cite{OfficeHome} is a much larger benchmark containing 65 different class images from 4 domains: Ar (Art), Cl (Clipart), Pr (Product), and Rw (RealWorld). Following the existing evaluation settings \cite{PADA,ETN}, we have 12 partial domain adaptation tasks. From each target domain, we only select the first 25 categories in alphabetical order, while the source domain utilizes all 65 class images.

\begin{table*}
\linespread{1.15} 
\centering
\caption{Comparisons of Recognition Rates ($\%$) of Partial Domain Adaptation on Office+Home Dataset (ResNet-50).} 
\vspace{-3mm}
\label{OfficeHome} 
\setlength{\tabcolsep}{3pt} 
\renewcommand{\arraystretch}{1} 
\begin{tabular}{c|cccccccccccc|c} 
\hline
Method & Ar $\rightarrow$ Cl & Ar$\rightarrow$Pr & Ar$\rightarrow$Rw & Cl$\rightarrow$Ar & Cl$\rightarrow$Pr & Cl$\rightarrow$Rw & Pr$\rightarrow$Ar & Pr$\rightarrow$Cl & Pr$\rightarrow$Rw & Rw$\rightarrow$Ar & Rw$\rightarrow$Cl & Rw$\rightarrow$Pr & Avg. \\   
\hline
Source Only & 46.33 & 67.51 & 75.87 & 59.14 & 59.94 & 62.73 & 58.22 & 41.79 & 74.88 & 67.40 & 48.18 & 74.17 & 61.35 \\ 
DAN \cite{DAN}& 43.76 & 67.90 & 77.47 & 63.73 & 58.99 & 67.59 & 56.84 & 37.07 & 76.37 & 69.15 & 44.30 & 77.48 & 61.72 \\
DANN \cite{DANN} & 45.23 & 68.79 & 79.21 & 64.56 & 60.01 & 68.29 & 57.56 & 38.89 & 77.45 & 70.28 & 45.23 & 78.32 & 62.82 \\
ADDA \cite{ADDA}& 45.23 & 68.79 & 79.21 & 64.56 & 60.01 & 68.29 & 57.56 & 38.89 & 77.45 & 70.28 & 45.23 & 78.32 & 62.82 \\
RTN \cite{RTN}& 49.31 & 57.70 & 80.07 & 63.54 & 63.47 & 73.38 & 65.11 & 41.73 & 75.32 & 63.18 & 43.57 & 80.50 & 63.07 \\
IWAN \cite{IWAN}& 53.94 & 54.45 & 78.12 & 61.31 & 47.95 & 63.32 & 54.17 & 52.02 & 81.28 & \underline{76.46} & 56.75 & 82.90 & 63.56 \\
SAN \cite{SAN} & 44.42 & 68.68 & 74.60 & \textbf{67.49} & 64.99 & \textbf{77.80} & 59.78 & 44.72 & 80.07 & 72.18 & 50.21 & 78.66 & 65.30 \\
PADA \cite{PADA}& 51.95 & 67.00 & 78.74 & 52.16 & 53.78 & 59.03 & 52.61 & 43.22 & 78.79 & 73.73 & 56.60 & 77.09 & 62.06 \\
DRCN \cite{DRCN} & 54.00 & 76.40 & 83.00 & 62.10 & 64.50 & 71.00 & \textbf{70.80} & 49.80 & 80.50 & \textbf{77.50} & 59.10 & 79.90 & 69.00 \\
ETN \cite{ETN} & 59.24 & 77.03 & 79.54 & 62.92 & 65.73 & 75.01 & \underline{68.29} & \underline{55.37} & \textbf{84.37} & 75.72 & 57.66 & \textbf{84.54} & 70.45 \\
\hline
Ours($C_N$) & \underline{61.41} & \underline{83.81} & \underline{86.36} & 64.15 & \underline{74.12} & \underline{75.15} & 67.22 & \textbf{55.44} & \underline{83.88} & 72.15 & \underline{60.22} & 83.59 & \underline{72.29}\\
Ours($C_P$) & \textbf{62.54} & \textbf{83.92} & \textbf{86.69} & \underline{65.44} & \textbf{74.96} & 75.04 & 67.40 & 55.14 & \textbf{84.37} & 73.25 & \textbf{60.51} & \underline{84.09} & \textbf{72.78}\\
\hline

\end{tabular}\vspace{-3mm}
\end{table*}


\begin{table*}[h]
\linespread{1.15} 
\centering
\caption{Comparisons of Recognition Rates ($\%$) of Unsupervised Domain Adaptation on Office+Home Dataset (ResNet-50).} 
\label{strategy} \vspace{-3mm}
\setlength{\tabcolsep}{2pt} 
\renewcommand{\arraystretch}{1} 
\begin{tabular}{c|c|cccccccccccc|c} 

\hline
\multicolumn{2}{c|}{Method} & Ar $\rightarrow$ Cl & Ar$\rightarrow$Pr & Ar$\rightarrow$Rw & Cl$\rightarrow$Ar & Cl$\rightarrow$Pr & Cl$\rightarrow$Rw & Pr$\rightarrow$Ar & Pr$\rightarrow$Cl & Pr$\rightarrow$Rw & Rw$\rightarrow$Ar & Rw$\rightarrow$Cl & Rw$\rightarrow$Pr & Avg. \\   
\hline
\multirow{2}{*}{No Adaptive} & $C_N$ & 51.79 & 70.42 & 79.40 & 56.16 & 62.97 & 70.40 & 60.42 & 48.15 & 76.75 & 66.08 & \textbf{63.94} & 76.58 & 65.26 \\
~& $C_P$ & 51.31 & 70.31 & 79.18 & 56.16 & 63.08 & 70.04 & 60.51 & 48.03 & 75.76 & 66.08 & 53.52 & 76.64 & 64.25 \\
\multirow{2}{*}{$C_N$ Guide} & $C_N$ & \underline{62.09} & 81.01 & 83.60 & 60.75 & 64.48 & 65.27 & 65.20 & 53.52 & \textbf{84.76} & 71.23 & 56.39 & 80.06 & 69.03 \\
~& $C_P$ & 61.95 & 80.84 & 83.32 & 60.94 & 64.71 & 65.93 & 65.56 & 53.58 & \textbf{84.76} & 71.14 & 56.39 & 79.89 & 69.08\\
\multirow{2}{*}{Same $C_N$\&$C_P$} & $C_N$ & 56.75 & 80.06 & \underline{87.36} & 60.20 & 64.99 & \textbf{76.97} & 65.75 & \underline{55.14} & 83.27 & 69.30 & 55.08 & 82.18 & 69.75\\
~& $C_P$ & 56.81 & 80.00 & \textbf{87.41} & 60.29 & 64.93 & \textbf{76.97} & 65.75 & 55.08 & 83.27 & 69.30 & 55.02 & 82.18 & 69.75\\
\hline
\multirow{2}{*}{Ours} & $C_N$ & 61.41 & \underline{83.81} & 86.36 & \underline{64.15} & \underline{74.12} & \underline{75.15} & \underline{67.22} & \textbf{55.44} & 83.88 & \underline{72.15} & 60.22 & \underline{83.59} & \underline{72.29}\\
& $C_P$ & \textbf{62.54} & \textbf{83.92} & 86.69 & \textbf{65.44} & \textbf{74.96} & 75.04 & \textbf{67.40} & \underline{55.14} & \underline{84.37} & \textbf{73.25} & \underline{60.51} & \textbf{84.09} & \textbf{72.78}\\
\hline

\hline

\end{tabular}\vspace{-3mm}
\end{table*}

\noindent\textbf{Comparisons}: We compare the performance of our proposed method with several domain adaptation and the state-of-the-art partial DA methods such as: Deep Adaptation Network (DAN) \cite{DAN}, Adversarial Discriminative Domain Adaptation (ADDA) \cite{ADDA}, Residual Transfer Network (RTN) \cite{RTN}, Importance Weighted Adversarial Nets (IWAN) \cite{IWAN}, Selective Adversarial Network (SAN) \cite{SAN}, Partial Adversarial Domain Adaptation (PADA) \cite{PADA}, Example Transfer Network (ETN) \cite{ETN}, and Adaptive Feature Norm (AFN) \cite{AFN}. Specifically, DAN applies multi-kernel MMD to match source and target domain distribution and learn transferable features across the domain. ADDA combines the adversarial training idea and united weights sharing to generate domain invariant features. RTN jointly adapts features distribution as well as source and target classifiers via deep residual learning framework. IWAN and SAN select or re-weight outlier categories in source domain label space to alleviate the negative influence caused by those classes that are not in the target domain label space. PADA, ETN, and AFN are the state-of-the-art partial domain adaptation models. Through down-weighting source domain data which is from outlier categories, PADA reduces the negative transfer influence caused by outlier classes. ETN proposes a progressive weighting scheme to quantify the transferability of source examples. AFN proposes a parameter-free approach to progressively adapt the source and target domain feature norms to a large range of values, which results in significant transfer gains.

\noindent\textbf{Implementation Details}: For each source-target pair case, we finetune the ImageNet pre-trained convolutional neural networks on the source domain and remove the last fully-connected layer as the backbone network. Then we input the backbone networks output of all source and target domain data into two dense layers with hidden layer output as 1,024 followed by ReLU activation and 0.1 dropout probability as the feature extractor $G(\cdot)$. We accept ResNet-50 network \cite{he2016deep} as the backbone on Office-Home and Office-31, and also explore the performance of VGG network as the backbone \cite{simonyan2014very} on Offce-Home dataset.
 The output dimension of the generator $G(\cdot)$, as known as the embedding features $\mathbf{z}_{x/t}$, is 512. The multilayer perceptron classifier $C_N(\cdot)$ is a two-layer fully-connected neural network where the hidden layer output dimension is 512, and the output size is the number of source domain categories. For prototype classifier $C_P(\cdot)$, we take cosine similarity as the measurement function in $\mathbf{\Phi}(\cdot,\cdot)$, and we directly take the source domain class centers as the prototypes, because the feature generator update every epoch, so the prototypes are also updating along with training. All experiments are implemented via PyTorch. We train the model for 100 epochs by Adam optimizer with learning rate as 0.0001, and report the last epoch results. $\mathbf{p}_0$ is rounded to two decimal places. $\lambda_1 = 0.1$ and $\lambda_2 = 0.5$ on Office31 dataset, while $\lambda_1=0.01$, $\lambda_2=2$ on Office-Home. We will analyze the parameter sensitivity in Section \ref{Analysis}.

\subsection{Comparison Results}

In this section, we will comprehensively evaluate our proposed model with several baselines on Office-31 and Office-Home benchmarks in terms of the target samples labels prediction accuracy to manifest the effectiveness of our model.

Specifically, we observe that
PDA methods (IWAN, SAN, PADA, DRCN, and ETN) achieve better performance than standard DA efforts such as DAN, DANN, ADDA, and RTN.
ETN achieves much greater improvement because it introduces a method to quantify the source samples transferability.
Our proposed method can still outperform all compared baselines on most partial domain adaptation tasks and obtain the best average performance.

Table \ref{Office_Res} reports the classification accuracy on the Office-31 dataset obtained by all baselines and our model with ResNet-50 as the backbone of feature extractor. It is noteworthy that the prototype classifier $C_P(\cdot)$ always generates better performance than the conventional multilayer perceptron classifier $C_N(\cdot)$. From the results, the prototype classifier achieves the best performance on 5 out of 6 tasks, compared to all the other baselines. To be specific, the average classification accuracy reaches the best performance $\mathbf{97.72\%}$, and reaches $\mathbf{100\%}$ accuracy on W31 $\rightarrow$ D10 and D31 $\rightarrow$ W10. 

Moreover, we also explore the VGG network as the feature extractor backbone on Office-31 dataset and report the results in table \ref{Office_VGG}. Our proposed model achieves the best average performance compared with other baselines. Specifically, compared to the best baseline performance on task A31$\rightarrow$W10, PADA, $C_N(\cdot)$ and $C_P(\cdot)$ improve the accuracy over 2\% to 88.44\% and 4\% to 90.48\%, respectively. It is noteworthy that the improvements of performance with VGG networks as backbone is more significant than using ResNet-50, because the ResNet-50 is more advanced deep convolutional neural networks model, which can generate more task specific discriminate features than VGG networks. 

Experiment results on the Office-Home dataset are stated in Table \ref{OfficeHome}. Both $C_N(\cdot)$ and $C_P(\cdot)$ obtain better performance against other baselines with significant improvements on average classification accuracy ($\mathbf{1.84\%}$ and $\mathbf{2.33\%}$). Moreover, our proposed method achieves more than $\mathbf{5\%}$ accuracy increase compared to the state-of-the-art baseline, e.g., Ar $\rightarrow$ Pr, Cl $\rightarrow$ Pr, etc. 

\begin{figure*}[h]
    \centering
    \includegraphics[width=1\linewidth]{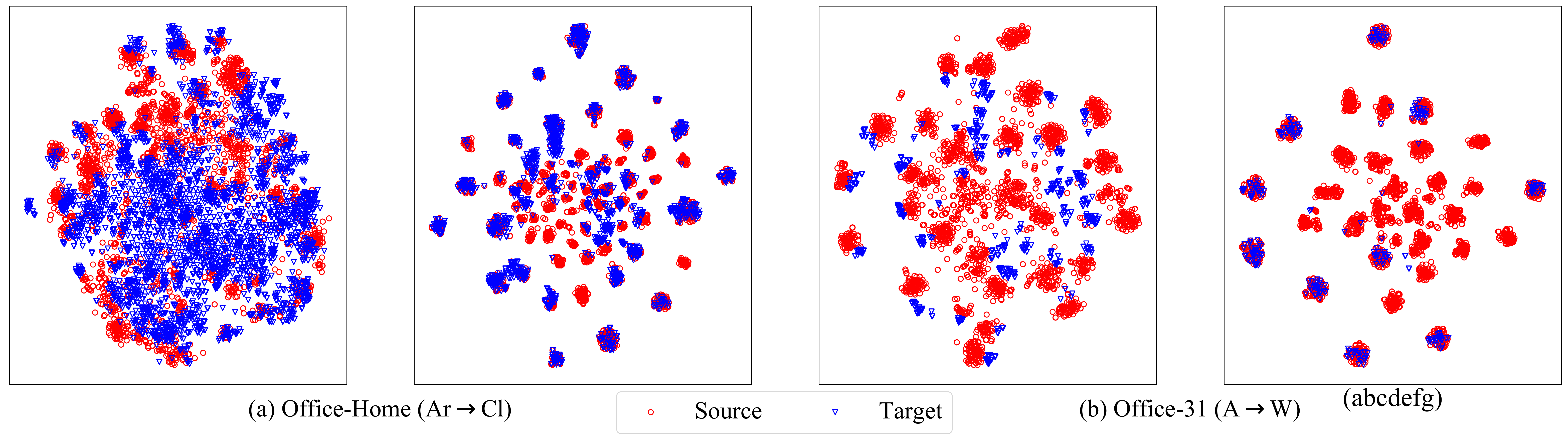}
    \vspace{-8mm}
    \caption{tSNE visualization of the original features and generator $G(\cdot)$ output embedding features after domain adaptation. (a) Office-Home dataset (Al $\rightarrow$ Cl) (b) Office-31 dataset (Amazon $\rightarrow$ Webcam). }\vspace{-3mm}
    \label{fig:tsne}
\end{figure*}

\begin{figure*}[h]
    \centering
    \includegraphics[width=1\linewidth]{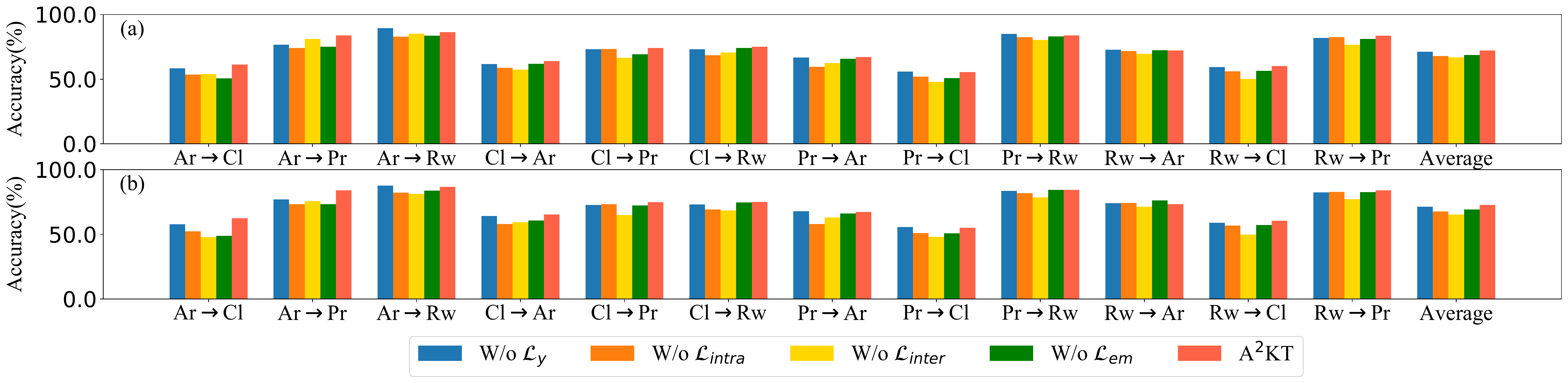}
    \vspace{-8mm}
    \caption{Evaluate each loss term contribution on Office-Home dataset by removing each specific term while keeping other parts same. (a) multilayer perceptron classifier $C_N(\cdot)$ (b) Prototype Classifier $C_P(\cdot)$. }
    \label{fig:ablation}\vspace{-4mm}
\end{figure*}

\subsection{Ablation Analysis}
\label{Analysis}


\begin{figure}
    \centering
    \includegraphics[width=1\linewidth]{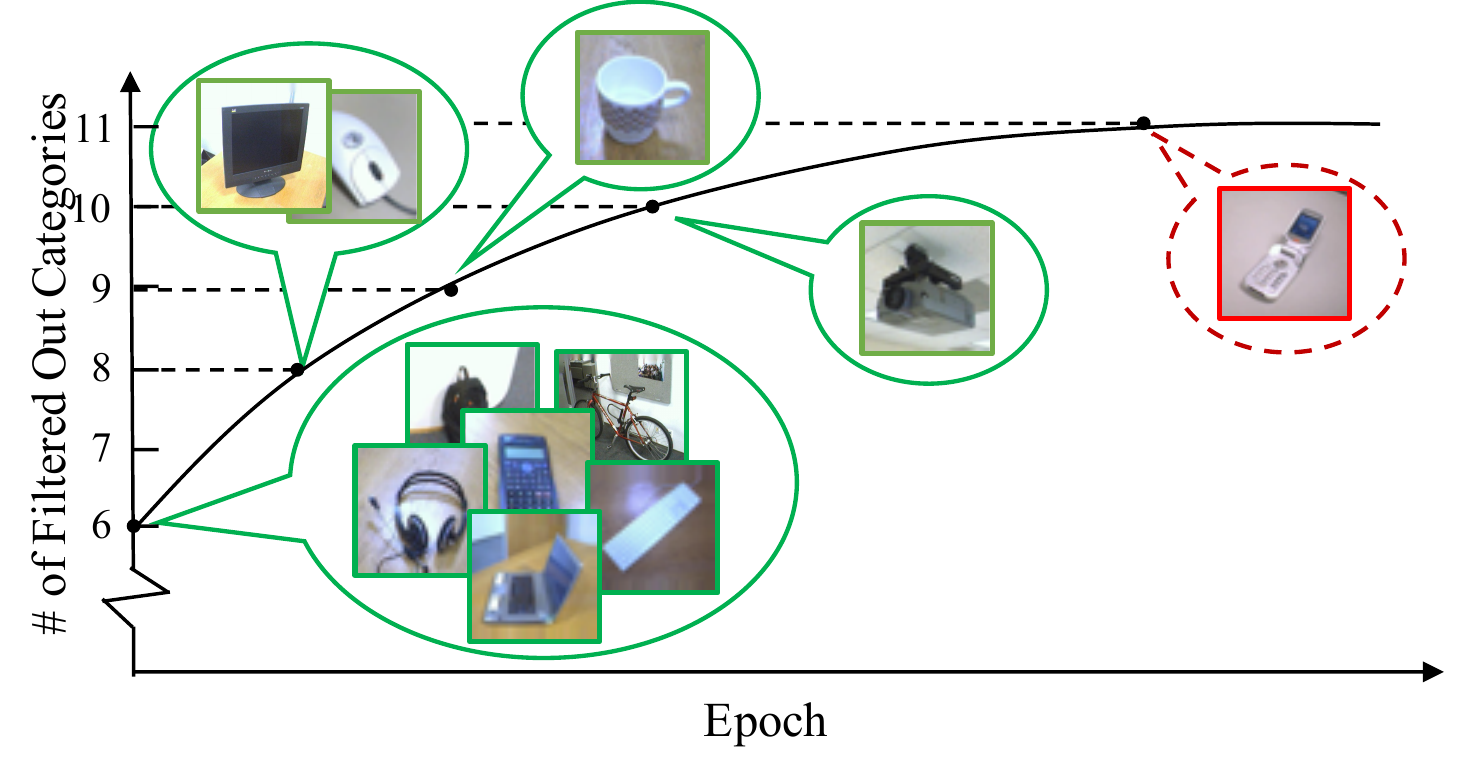}
    \vspace{-8mm}
    \caption{Filtered out shared categories of the target domain of task A31 $\rightarrow$ W10 on the Office-31 dataset.}\vspace{-6mm}
    \label{fig:adaptation}
\end{figure}

First, visualize the generator $G(\cdot)$ output features  before and after the domain adaptation process on task Ar$\rightarrow$Cl on Office-Home, and A$\rightarrow$W on Office-31 dataset in the Fig. \ref{fig:tsne} (a) and (b). From the results, we observe that our proposed method  aligns the source and target domain samples with respect to categories, and tights the compactness of the embedding features to each class centers.

Secondly, we evaluate the contribution of every loss term in Eq. \eqref{opt_overall} by removing each specific term while keeping other terms as the original framework. The results are shown in Fig. \ref{fig:ablation}. It is noteworthy that both $\mathcal{L}_{intra}$ and $\mathcal{L}_{inter}$ make crucial contribution to the PDA tasks because these two terms are aligning the data distribution inter-classes and intra-class. $\mathcal{L}_y$ keeps the model performance on the source domain stable, while it has limited contribution to the PDA process, but cannot be ignored. $\mathcal{L}_{em}$ helps to mitigate the negative transfer influence of the multilayer perceptron classifier $C_N$, especially at the beginning of the training stage.

Then, we monitor the training and optimization process of our model. Fig. \ref{fig:adaptation} illustrates the process of the \textit{adaptively-accumulated knowledge transfer} process. We choose case A31 $\rightarrow$ W10 of Office-31 dataset and show the changing of the filtered out high prediction confidence categories used to align the data distribution across domains. In the beginning, high prediction target samples only spread in only 6 classes, but then more and more categories are involved, and the number finally reaches 11, while the total number of the target domain categories is 10. Although there is an incorrect outlier class involved, the adaptive optimization strategy still significantly narrows the range of the target domain label space.

\begin{figure}
    \centering
    \includegraphics[width=1\linewidth]{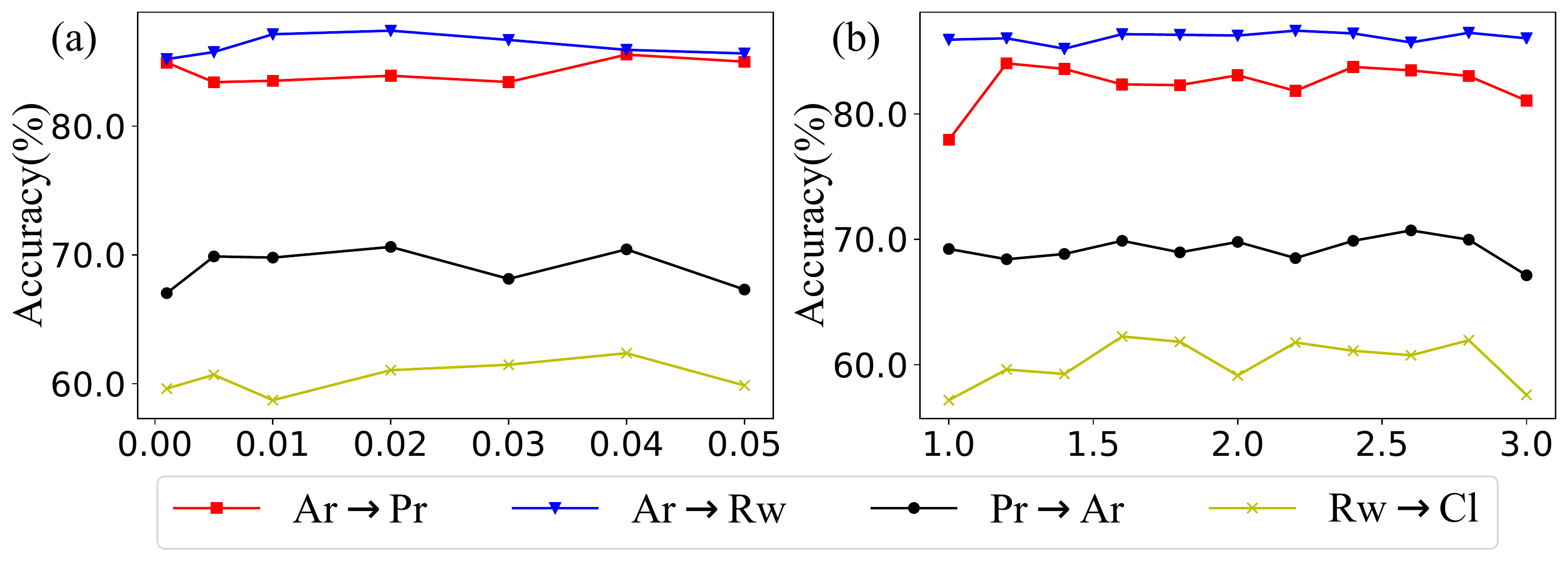}\vspace{-2.5mm}
    \caption{Parameters sensitivity analysis of (a) $\lambda_1$  (b) $\lambda_2$ on 4 different tasks from Office-Home dataset.}
    \label{fig:parameters}\vspace{-2mm}
\end{figure}

\begin{figure*}
    \centering
    \includegraphics[width=1\linewidth]{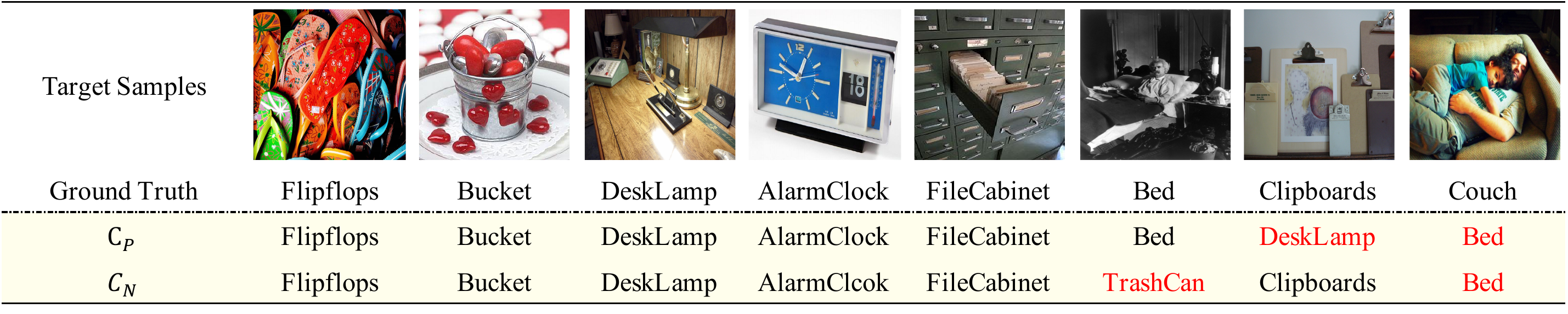}
    \vspace{-7mm}
    \caption{Prediction of $C_N(\cdot)$ and $C_P(\cdot)$ for selected target domain samples (Pr $\rightarrow$ Rw) }
    \label{fig:top}\vspace{-3mm}
\end{figure*}

\begin{figure*}
    \centering
    \includegraphics[width=1\linewidth]{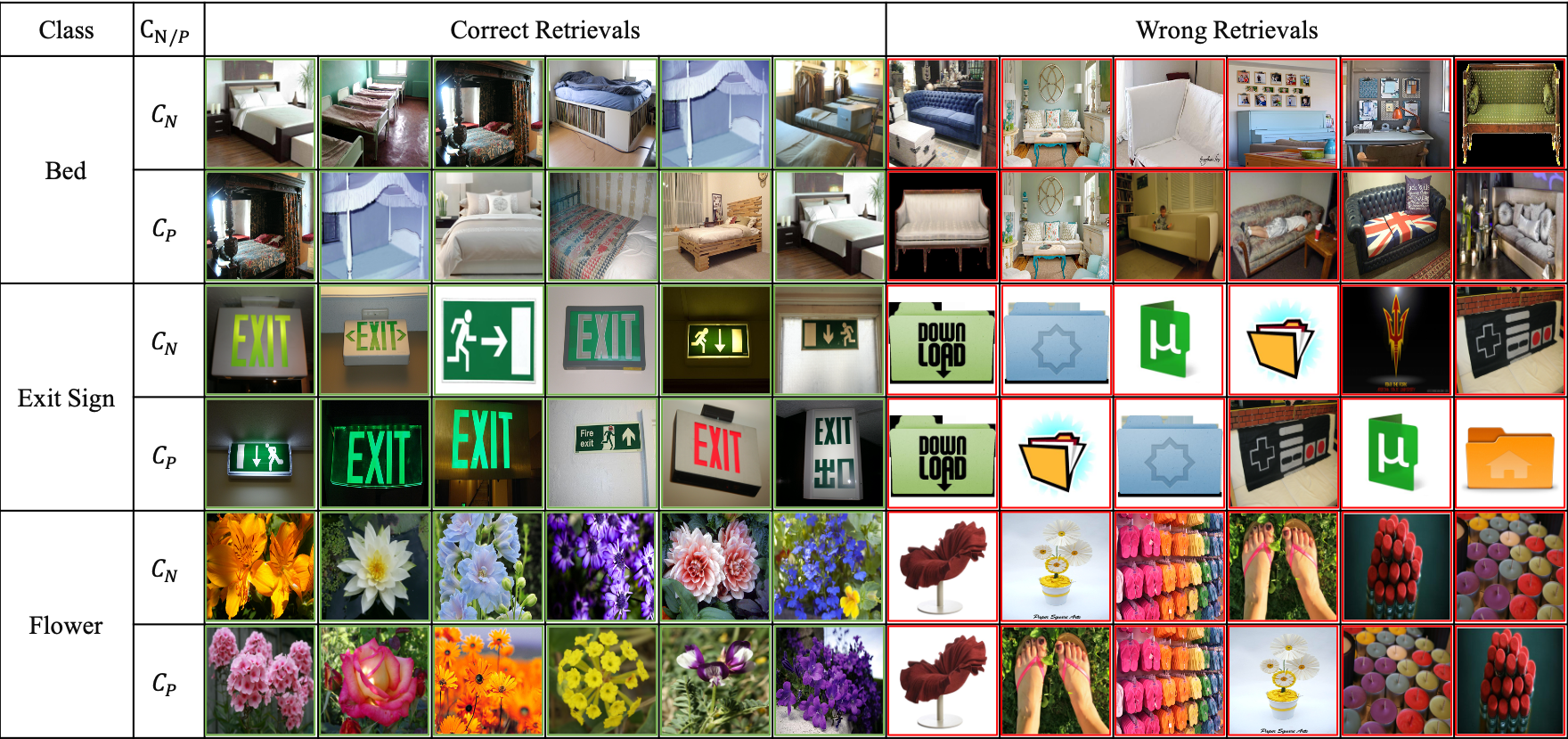}
    \vspace{-7mm}
    \caption{Retrieved target images with the highest 10 prediction confidence by $C_P(\cdot)$ (Pr $\rightarrow$ Rw).}
    \label{fig:retrieval}\vspace{-3mm}
\end{figure*}

Moreover, we implement several ablation experiments on the Office-Home dataset with different training details to explore the contribution of our proposed model and optimization strategy, the results are reported in Table \ref{strategy}. "No Adaptive" denotes the results without the adaptively accumulating knowledge transfer and target samples filtering out process. From the results, compared to our complete A$^2$KT model results, we notice how important the \textit{adaptively accumulating knowledge} strategy is. "$C_N$ Guide" are the results when we use the $C_N$ probabilistic prediction to filter out high confidence target samples for domain alignment, instead of $C_P$. The way to decide the threshold is the same as when we use $C_P$. The results prove that the multilayer perceptron classifier $C_N$ and the prototype classifier $C_P$ have different classification philosophy, and using $C_P$ probability prediction to accumulate can boost the performance significantly. Finally, we explore the motivation of adopting two different type dual classifiers framework in our model by setting $C_N$ and $C_P$ both same structure multilayer perceptron classifiers, all other settings and training strategies are the same as before, and the results are reported in "Same $C_N \& C_P$". From the results, we observe that for some cases two same multilayer perceptron classifiers can get slightly better performance than our model, e.g., Ar $\rightarrow$ Rw and Cl $\rightarrow$ Rw. However, for most cases and the average performance, our model with different type classifiers outperforms much more. All the results with different training strategies in Table \ref{strategy} demonstrate the effectiveness and motivation of our model and optimization strategies.

We present the parameter sensitivity analysis in Fig. \ref{fig:parameters}. We vary $\lambda_1$ from 0.0001 to 0.05 and $\lambda_2$ from 1 to 3 on four cases on the Office-Home dataset (Ar $\rightarrow$ Pr, Ar $\rightarrow$ Rw, Pr $\rightarrow$ Ar, Rw $\rightarrow$ Cl) to analyze if the model is sensitive to the change of the hyper-parameters. The results in Fig. \ref{fig:parameters} shows that our model has great stability across cases of the two parameters $\lambda_1$ and $\lambda_2$.

Finally, we select several representative target samples from task Pr$\rightarrow$Rw on Office-Home dataset and show the predictions of $C_N(\cdot)$ and $C_P(\cdot)$ in Fig. \ref{fig:top}. We notice that 
some cases only $C_N(\cdot)$ or $C_P(\cdot)$ can handle, or even neither can predict correctly, which demonstrates the motivation of combine two different type classifiers $C_N(\cdot)$ and $C_P(\cdot)$ in our proposed model. Besides, we operate the image retrieval task by giving specific labels to retrieve the target samples. The 5 target images with the highest $C_P(\cdot)$ prediction confidence and 5 with the lowest in the retrieved images are shown in Fig. \ref{fig:retrieval}. The different samples retrieved by $C_N(\cdot)$ and $C_P(\cdot)$ demonstrate the motivation of integrating various classifiers. 

\section{Conclusion}
This paper presented a novel Domain-Invariant Feature learning framework for partial domain adaptation. With the help of the Adaptively-Accumulated Knowledge Transfer Optimization strategy, the target domain samples with high confidence and task-relevant source categories are selected out adaptively. By maximizing the inter-class center-wise discrepancy and minimizing the intra-class sample-wise compactness, more domain-invariant and task-specific discriminative representations will be extracted. Extensive experiments on several partial domain adaptation benchmarks manifest the superiority of our algorithms against previous works.
\clearpage
\bibliographystyle{ACM-Reference-Format}
\bibliography{sigconf}

\end{document}